# Unified Class and Domain Incremental Learning with Mixture of Experts for Indoor Localization

Akhil Singampalli
Department of Electrical and Computer Engineering
Colorado State University
Fort Collins, CO, USA
Akhil.Singampalli@colostate.edu

Sudeep Pasricha
Department of Electrical and Computer Engineering
Colorado State University
Fort Collins, CO, USA
Sudeep@colostate.edu

*Abstract*— Indoor localization using machine learning has gained traction due to the growing demand for location-based services. However, its long-term reliability is hindered by hardware/software variations across mobile devices, which shift the model's input distribution to create domain shifts. Further, evolving indoor environments can introduce new locations over time, expanding the output space to create class shifts, making static machine learning models ineffective over time. To address these challenges, we propose a novel unified continual learning framework for indoor localization called MOELO that, for the first time, jointly addresses domain-incremental and class-incremental learning scenarios. MOELO enables a lightweight, robust, and adaptive localization solution that can be deployed on resource-limited mobile devices and is capable of continual learning in dynamic, heterogeneous real-world settings. This is made possible by a mixture-of-experts architecture, where experts are incrementally trained per region and selected through an equiangular tight frame based gating mechanism ensuring efficient routing, and low-latency inference, all within a compact model footprint. Experimental evaluations show that MOELO achieves improvements of up to 25.6× in mean localization error, 44.5× in worst-case localization error, and 21.5× lesser forgetting compared to state-of-the-art frameworks across diverse buildings, mobile devices, and learning scenarios.

## I. Introduction

Indoor localization refers to the estimation of a device or user's position within enclosed environments. Unlike outdoor positioning, which relies on GPS/GNSS, indoor localization is challenging due to the absence of satellite signals and the complexity of indoor structures. Accurate indoor positioning is driving the location-based services (LBS) market from USD 31.17 billion in 2024 to a projected USD 125.92 billion by 2032 [1]. This growth is fueled by the rise of smartphones, accelerating deployment of IoT ecosystems, and emerging AR/VR applications [2]. Among the technologies enabling this shift, Wi-Fi-based localization offers a practical solution due to its widespread availability indoors, low deployment cost, and seamless integration with existing infrastructure [3].

One widely used method for Wi-Fi-based indoor localization is fingerprinting [4]. This approach typically involves two phases: an offline phase and an online phase. In the offline phase, Wi-Fi Received Signal Strength (RSS) measurements are collected at fixed locations (known as Reference Points or RPs) to build a fingerprint database. A machine learning (ML) model is then trained to map these fingerprints to known positions. In the online phase, the model uses real-time RSS measurements to estimate the user's location based on the learned mappings [5].

A major challenge with Wi-Fi fingerprinting-based indoor localization is that indoor environments are inherently dynamic. RSS values fluctuate naturally due to factors like multipath effects, human mobility, and temporal variations such as changes in occupancy or environmental conditions [4]. In addition, device heterogeneity (the fact that different smartphones sense RSS differently) creates further variability [6]. Over time, physical changes such as furniture rearrangements and renovations alter the signal propagation landscape. Together, these variations lead to a *domain shift*, where the statistical distribution of RSS signals in deployment differs from that used during training. In parallel, deployments also face *class shift*: the set of indoor locations (RPs) evolves as new regions are added, expanding the label space and requiring models to incorporate new classes in its output. This mismatch degrades model accuracy, forcing updates to both known RPs and new RPs, from previously unseen areas.

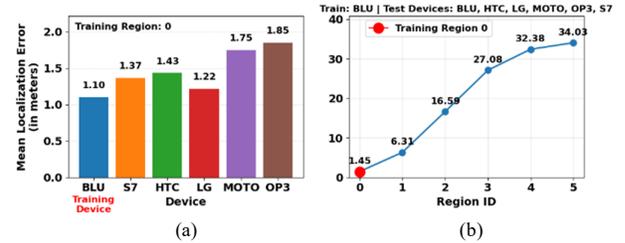

Fig. 1. Illustration on impact of (a) domain shifts and (b) class shifts on indoor localization performance (localization error in meters).

To quantify the impact of these changes post-deployment, we conducted an experiment using the state-of-the-art FEDHIL indoor localization framework [7] in a real building. The ML model in FEDHIL is initially trained on data collected from 10 RPs (locations) in region ID 0 using a BLU Vivo 8 (BLU) mobile device and evaluated on the BLU, Samsung Galaxy S7 (S7), HTC U11 (HTC), LG V20 (LG), Motorola Z2 (MOTO), OnePlus 3 (OP3) mobile devices. As shown in Fig. 1(a), FEDHIL on the same trained indoor region (ID 0) but running across different devices (x-axis) leads to notably higher localization errors (y-axis) due to device heterogeneity, indicating a *domain shift*. Fig. 1(b) further demonstrates that FEDHIL running on new indoor regions (IDs 1-5) with new RPs distant from the training region results in sharply rising localization errors, pointing to a *class shift*. Together, these results emphasize the need for ML frameworks that can adapt continuously to both new devices and new regions over time.

Continual Learning (CL) offers a compelling path forward for building adaptive indoor localization systems that remain effective over time [4]. A key challenge that must be addressed with CL is catastrophic forgetting, where models lose previously learned knowledge when adapting to new data. This becomes critical in two common scenarios: domain-incremental learning (DIL), where the model must adapt to shifts in input distributions caused by changes in devices and environments, and class-incremental learning (CIL), where new RPs are added, expanding the output space. Most existing solutions treat these challenges separately, limiting their practical use. However, real-world deployments demand a unified approach that can handle both simultaneously.

To this end, we introduce MOELO, a unified continual learning framework designed to jointly enable DIL and CIL. MOELO supports efficient, incremental updates without retraining from scratch, enabling robust and scalable localization in dynamic, heterogeneous environments. The novel contributions of this work are:

- A novel unified framework for indoor localization that, for the first time, jointly addresses both DIL and CIL, enabling robust adaptation to dynamic real-world environment.
- A customized gated mixture-of-experts ML architecture that dynamically allocates experts to regions while sharing knowledge across devices.
- An Equiangular Tight Frame (ETF)-based gating mechanism that ensures stable expert selection, improving scalability with the number of regions.
- A device-region balanced replay strategy that maintains a compact memory while preserving coverage across both devices and spatial regions.
- Extensive evaluations on real-world data across diverse buildings, mobile devices, and scenarios reflecting realistic domain and class shifts in indoor locales.

## II. RELATED WORKS

Research on Wi-Fi fingerprinting has been driven by competitions, notably the IPIN series [8] and Microsoft-sponsored challenges [9], which emphasize practical concerns like smartphone heterogeneity, temporal drift, and on-device constraints. These efforts have highlighted many challenges for building robust indoor localization systems that generalize across diverse real-world conditions. Early ML-based approaches used k-nearest neighbors (KNN) [10], hidden Markov models (HMM) [11], and Gaussian process classifiers (GPC) [12] to address short-term RSS variability due to multipath effects and human activity but generally lacked the representational capacity to generalize across diverse devices. Subsequent ML-based framework such as DNNLOC [13], CNNLOC [14], ANVIL [15], VITAL [16] and STELLAR [17] improved accuracy by learning discriminative device features and attention mechanisms, yet they are typically trained once and deployed statically, leaving them sensitive to post-deployment shifts in devices and environmental factors.

Federated learning (FL) has been applied to indoor localization to enable online adaptation post-deployment without sharing raw RSS data. Most FL methods use standard aggregation techniques like FedAvg or adjust for device differences by weighting client updates to reduce the impact of noisy data [18]. FedLoc [19] and FEDHIL [7] show that FL can improve localization accuracy while preserving privacy. However, these methods often assume a fixed set of locations and struggle to support new RPs over time. Approaches that rely on pseudo-labels, e.g., [20], under domain shift can also introduce errors, leading to unstable global models.

Another line of work focuses on handling domain shift in indoor localization. Domain incremental learning (DIL) methods aim to make localization more robust across devices and environments by learning domain-invariant features. Techniques include reconstruction-based learning using VAEs [21], adversarial training [22], distribution alignment [23], Siamese networks for comparing signal patterns [24], and memory-guided class latent alignment [25]. These methods can work well when adapting from one known source to one target domain. However, they are usually designed for one-time adaptation and lack the ability to continuously learn as new devices or RPs (locations) are introduced over time.

Class-incremental learning (CIL) methods have been explored to improve long-term adaptability of neural networks [26]. Common strategies include regularizing important model parameters such as EWC [27], [28], expanding the network architecture as in PNN [29], or replaying past examples through stored data [30]. While these techniques help reduce forgetting, most of these rely on predefined learning phases and high resource availability conditions rarely met in real-world indoor localization, where updates must be lightweight and efficient. A recent work CIELO [31] implements CIL for indoor localization by expanding the RP label space with memory replay and sample selection, but it does not address domain shifts across devices and environments. Beyond indoor localization, MoCo-ON [32] addresses both CIL and DIL with an orthogonal negative contrastive loss that improves generalization across large domain gaps. But our analysis (Section V) shows that the framework struggles to remain stable in indoor localization systems that involve dynamic class shifts.

*In summary,* there remains a crucial need for a unified and practical continual learning solution for indoor localization. MOELO addresses this need through a mixture-of-experts that specializes across devices and regions, maintaining a compact trainable footprint and enabling selective adaptation. To the best of our knowledge, MOELO is the first framework to support both CIL and DIL for indoor localization.

## III. INCREMENTAL LEARNING IN INDOOR LOCALIZATION

ML-based indoor localization systems must remain accurate over time, even as conditions evolve. In real deployments, this includes the addition of new spatial regions, the use of different mobile devices, and environmental changes such as signal interference. These dynamics disrupt the assumptions of static ML models by altering both the input distribution and output space. Addressing such variability requires a CL approach, where the system adapts to new conditions and regions while mitigating forgetting, preserving previously learned knowledge even as it gradually incorporates new knowledge over time.

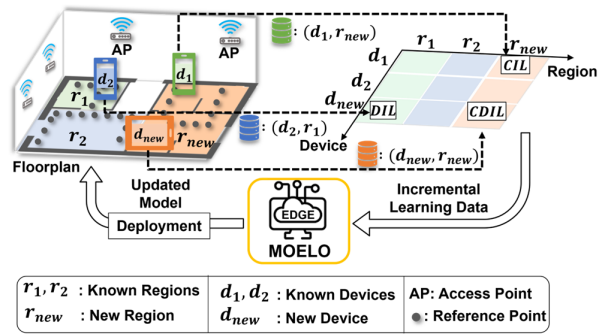

Fig. 2. Overview of a CL-based indoor localization framework

Fig. 2 illustrates the CL setup for indoor localization. The floorplan space on the left is divided into multiple regions ($r_1$, $r_2$,…, $r_{new}$), and Wi-Fi RSS fingerprints are gathered as users move through these areas with different mobile devices ($d_1$, $d_2$,…, $d_{new}$). Each sample is labeled with a tuple of device ($d_i$) and region ($r_i$) i.e., ($d_i$, $r_i$) that records both the device type and the region of collection. Using this metadata, three types of learning scenarios can be distinguished:

- *Domain-Incremental Learning (DIL)*: Occurs when data is collected in a known region using a new device. For example, if $r_1$ was previously mapped using $d_1$ and $d_2$, then introducing new device $d_{new}$ in $r_1$ creates a domain shift. Here region labels (i.e., output space) remain unchanged, but the input distribution differs because $d_{new}$ has different captured Wi-Fi signal characteristics.
- *Class-Incremental Learning (CIL)*: Occurs when a new region is added from a previously seen device. For instance, if the ML model was trained on $r_1$ and $r_2$ then the introduction of new region $r_{new}$ from known device $d_1$ represents a class shift. The ML model must learn to recognize $r_{new}$ as an additional output class.
- *Class and Domain -Incremental Learning (CDIL)*: Here, both a new device and a new region appear simultaneously, e.g., when a user carrying a new device $d_{new}$ collects data in a new region $r_{new}$. This case introduces the dual challenge of handling domain variation and expanding the output space at once.

These scenarios expose a fundamental learning challenge: catastrophic forgetting. As the model learns from new data, previously acquired knowledge may degrade, leading to poor localization in regions or conditions it once handled well. Addressing this requires a unified CL solution that can accommodate both domain and class shifts incrementally without costly retraining or sacrificing prior performance.

## IV. MOELO FRAMEWORK

MOELO is a modular framework built on a Mixture-of-Experts (MoE) design, providing a novel unified solution to both DIL and CIL for indoor localization. It adapts to mobile device heterogeneity, environmental variations, and the need to expand spatial coverage while preserving past knowledge. As shown in Fig. 3, MOELO processes Wi-Fi RSS fingerprints using a shared encoder to produces device-invariant representations, which are then routed to region-specific expert models. A lightweight Equiangular Tight Frame (ETF) based gating network assigns each fingerprint to the most relevant region expert, enabling accurate localization even under domain shifts while keeping computation and latency low. To handle class shift, MOELO introduces new experts for the new regions that help maintain performance on previously learned regions. During inference, only one expert is activated per fingerprint, which minimizes resource usage and makes the framework well-suited for smartphone deployment. MOELO further employs tailored training strategies and balanced replay to retain past knowledge while adapting to new conditions, as discussed next.

### A. MoE Architecture

At MOELO's core is a shared encoder that maps each Wi-Fi RSS fingerprint $x$ from incremental learning data to a latent representation $z$. The shared encoder is optimized to produce domain-invariant latent representations, so that samples from different devices but the same region cluster together in latent space. This is achieved with an Equiangular Tight Frame (ETF) based alignment using dot-regression loss, given as:

$$\mathcal{L}_{DR} = \frac{1}{2N}\sum_{i=1}^{N}\left(\cos(\hat{z}_i, v_{r_i}) - \frac{1}{\sqrt{R_{max}-1}}\right)^2 \quad (1)$$

$$where, \quad \cos(\hat{z}_i, v_{r_i}) = \frac{\langle \tilde{z}_i, \ v_{r_i}\rangle}{\|\ \tilde{z}_i\ \|_2\ \|v_{r_i}\|_2}$$

where, $N$ is batch size, $\hat{z}$ is the $L_2$-normalized projection of $z$, $v_{r_i}$ is the ETF anchor for region $r$ and sample $i$, and $R_{max}$ is total number of regions in a floorplan. The ETF anchors $W_{ETF} = \{v_r\}_{r=1}^{R_{max}}$ are $L_2$-normalized, uniformly spaced directions in latent space which are generated once based on a pre-defined $R_{max}$, determined by desired region granularity. As new regions appear, new experts are instantiated and bound to unused anchors, ensuring the routing geometry remains stable. $\mathcal{L}_{DR}$ in Eq. (1) measures the gap between each sample's cosine alignment to its region's anchor $v_r$ and the target equiangular tight frame (i.e., $1/\sqrt{K-1}$). Because a region shares one fixed anchor across domains, minimizing $\mathcal{L}_{DR}$ collapses domain variation within a region while maintaining angular separation between regions.

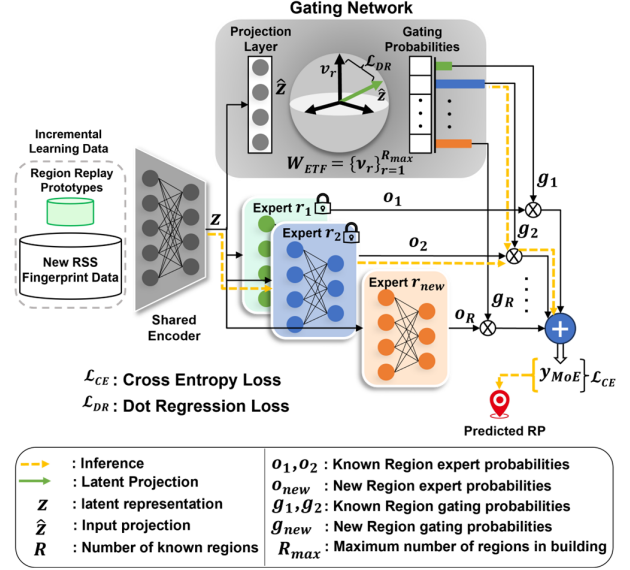

Fig. 3. Mixture-of-experts architecture for unified class and domain incremental learning in indoor localization.

### 1) Gating of Experts

The same ETF anchors define a simple, stable gating rule. With $L_2$-normalized projection layer output $\hat{z}$ and region anchor $v_r$, gating probabilities are given by:

$$g = softmax\left(\cos(\hat{z}, v_r)\right) \quad (2)$$

where $g$ outputs a probability distribution over all region experts, with each value reflecting how well the input projection $\hat{z}$ aligns with a region anchor. This serves as a routing signal to determine which experts should contribute to processing a given fingerprint and to what extent. During training, soft gating is employed allowing all experts to participate proportionally according to their weights in $g$, which encourages knowledge sharing and stable optimization. During inference, hard gating is applied to select only the expert whose anchor has highest similarity (highest probability) with $\hat{z}$. Fig. 3 illustrates this hard gating with the yellow dotted line. This step ensures efficient and specialized expert assignment, while keeping inference overheads low.

### 2) Region Experts

MOELO maintains $R$ experts, one per region encountered so far, and expands this set incrementally as new regions appear. For an input fingerprint $x$, each region expert receives the shared latent $z$ and outputs region probabilities ($o_r$) as:

$$o_r(z) = h_{\phi_r}(z) \in \mathbb{R}^{|c_r|} \quad (3)$$

where, $h_{\phi_r}$ is the expert network of region $r$ with parameters $\phi_r$ and $|\ c_r\ | = n_{RP}$ is the number of regional RP's handled by the expert. Let $c_r \subseteq C$ denote the local class set for region

$r$ in a global class space $C = \bigcup_{r=1}^{R} c_r$. To train on the global class space, each expert's probabilities $o_r(z)$ are first weighted by its gate weight $g_r$ (from Eq. (2)) and then concatenated across experts to form the fused probabilities ($y_{MoE}$) as shown:

$$y_{MoE}(x) = concat[g_1 o_1(z), g_2 o_2(z), \ldots, g_R o_R(z)] \quad (4)$$

The fused probabilities $y_{MoE}(x)$ span the entire global class set, allowing a single cross-entropy loss to be applied across all regions. This loss not only enforces discrimination among RPs (locations) within each indoor region but also ensures consistent separation across regions, effectively aligning the outputs of multiple experts into one unified classifier. Cross entropy loss for a batch of size $N$ is given by:

$$\mathcal{L}_{CE} = -\frac{1}{N} \sum_{i=1}^{N} y_i \log(y_{MoE,i}) \quad (5)$$

where, $y_i \in C$ is the true global class label of fingerprint $x_i$. Because each expert contributes only to its local class ($c_r$), experts specialize without interfering with each other.

### B. Continual Learning Strategy

MOELO jointly optimizes total loss as follows:

$$\mathcal{L}_{total} = \mathcal{L}_{CE} + \mathcal{L}_{DR} \quad (6)$$

This modular setup enables three distinct modes of continual learning:

- *DIL Training*: For new devices that appear in known indoor regions, only the encoder and gating projection are updated while freezing all experts. $\mathcal{L}_{DR}$ aligns new-device embeddings to existing anchors, removing device bias, while $\mathcal{L}_{CE}$ preserves class separation without disturbing decision boundaries.
- *CIL Training*: When a new region appears, a new expert is added and linked to the next ETF anchor, while keeping all prior experts frozen. Training then optimizes only $\mathcal{L}_{CE}$ on the new region class split $c_r$ so that weight updates are confined to the new expert and do not alter prior experts preventing forgetting. The equiangular spacing of anchors provide a stable routing geometry and prevents interfering with previously learned regions.
- *CDIL Training*: For a new region observed on new devices, a new expert (as in CIL) is added. The encoder along with gating (as in DIL) are updated, while keeping earlier experts frozen. Fixed anchors preserve old geometry; $\mathcal{L}_{DR}$ aligns new-device features to their anchors, and $\mathcal{L}_{CE}$ learns the new region achieving adaptation and expansion simultaneously.

### C. Region Replay Prototypes

To support long-term retention and continual adaptation, MOELO maintains a compact, device-region balanced replay memory. Each entry stores a Wi-Fi RSS fingerprint, label, device and region. As new data arrives, the memory buffer is updated using a herding-based strategy: for each (device, region) pair, fingerprint samples are evaluated using the current encoder, and a single representative fingerprint is selected that best approximates the mean embedding. The replay buffer allocates entries uniformly across pairs to ensure balanced representation. During training, each mini-batch includes both new fingerprints and region replay prototypes, enabling rehearsal that reinforces prior knowledge while adapting to new conditions.

This mechanism complements MOELO's continual learning strategy. Under DIL, replay anchors the encoder by revisiting past device distributions, preventing drift while enabling adaptation. In CIL, where prior experts are frozen, replay reinforces routing consistency by reminding the gating network of previously learned regions. In CDIL, replay plays a dual role: preserving alignment with old devices while also stabilizing routing as new experts are introduced. This memory-guided rehearsal when combined with frozen experts and fixed ETF anchors, allows MOELO to incrementally evolve without forgetting, achieving robust and scalable continual localization without full retraining.

## V. EXPERIMENTAL RESULTS

### A. Experimental Setup

We evaluate MOELO in a realistic continual learning setting with considerations for region expansion, device heterogeneity, and environment-driven signal drift.

To evaluate region expansion, two distinct buildings with varied layouts and construction materials are chosen. Each building features different numbers of reference points (RPs, i.e., indoor locations) and visible Wi-Fi access points (APs): Building 1 includes 60 RPs and up to 172 APs; Building 2 includes 48 RPs and 168 APs. All RPs were spaced at 1-meter intervals to enable fine-grained localization. To capture domain shifts caused by device heterogeneity, Wi-Fi RSS fingerprint data was collected using six different mobile devices with different hardware chipsets and software stacks, as shown in Table I. To evaluate environment-driven signal drift, fingerprints were collected across different time instances in the two buildings. The last column in Table I indicates how new devices were introduced at different time instances over a 3 month span during which variations in the indoor environment and Wi-Fi transceivers led to signal drift, providing a realistic post-deployment evaluation.

TABLE I. DETAILS OF MOBILE DEVICES USED IN EXPERIMENTS

| Manufacturer | Model | Acronym | Wi-Fi Chipset | Data Collection Instance |
|---|---|---|---|---|
| BLU | Vivo 8 | BLU | MediaTek Helio P10 | 0 |
| HTC | U11 | HTC | Qualcomm Snapdragon 835 | 9 hours |
| LG | V20 | LG | Qualcomm Snapdragon 820 | 1 Day |
| Motorola | Z2 | MOTO | Qualcomm Snapdragon 820 | 1 week |
| OnePlus | 3 | OP3 | Qualcomm Snapdragon 820 | 1 month |
| Samsung | Galaxy S7 | S7 | Qualcomm Snapdragon 835 | 3 months |

MOELO uses the MoE model composed of a shared encoder that maps Wi-Fi RSS vectors to a 64-dimensional latent space using two dense layers of 128 and 64 units with ReLU activations. The ETF gating network consists of a single projection layer that maps the 64-dimensional latent input to a 64-dimensional vector space, followed by L2 normalization. Each expert is a lightweight feedforward classifier with a single hidden layer of 128 units and dropout regularization, followed by an output layer sized to the subset of regional RPs it handles ($n_{RP}$). The granularity of regional partitioning is fixed to be $n_{RP} = 10$ for the experiments.

MOELO is compared against CIELO [31], which implements CIL through replay memory, MoCo-ON [32], which extends to both class and domain shifts by improving generalization under domain gaps, and DAILOC [25], which targets DIL through disentangled representations of device- and location-specific factors. These frameworks capture complementary aspects of class and domain shifts, making them strong baselines to evaluate against our framework.

TABLE II. COMBINATIONS OF CONTINUAL LEARNING SCENARIOS

| Incremental learning | Known region | New region |
|---|---|---|
| Known device | Baseline model | **CIL-Exclusive** |
| New device | **DIL-Exclusive** | **CDIL** |

Experiments were structured to evaluate three incremental learning scenarios as summarized in Table II, where combinations of known/new device and known/new regions define the different continual learning scenarios: CIL-Exclusive, DIL-Exclusive, and CDIL. Fig. 4 shows details of these scenarios, considering Building 1 as an example. Here, the DIL-Exclusive (DIL1–DIL6) scenario captures domain shifts with domain-incremental data collected across all RPs within Building 1 with new devices over time (as also shown earlier in Table I). The CIL-Exclusive (CIL1–CIL6) scenario models class shifts with class-incremental data collected from six mobile devices in the same region. Lastly, the CDIL (CDIL1–CDIL6) scenario represents the most challenging case, where class and domain shifts occur simultaneously. This structured design enables systematic evaluation of model robustness to both domain shifts and class shifts.

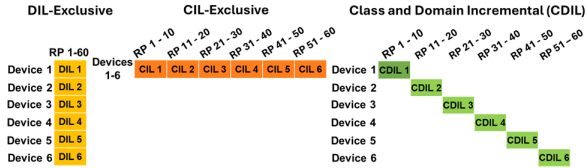

Fig. 4. Continual learning scenarios for Building 1.

The performance of MOELO and comparison frameworks was assessed using two primary metrics:

1) *Localization Error (LE)*: measured using the Euclidean distance between the predicted location $(X_{pred}, Y_{pred}, Z_{pred})$ and the ground truth location $(X_{true}, Y_{true}, Z_{true})$ given as:

$$LE = \sqrt{(X_{pred} - X_{true})^2 + (Y_{pred} - Y_{true})^2 + (Z_{pred} - Z_{true})^2} \quad (7)$$

providing a clear measure of indoor localization accuracy, where lower values correspond to better performance.

2) *Average Forgetting (AF)*: to measure the stability of continual learning, forgetting for an incremental learning unit $u$ (device or region) is defined as the difference between the units current localization error and its best localization error (achieved at any point during learning) expressed as:

$$Forgetting(F_u) = LE_u^{current} - LE_u^{best} \quad (8)$$

Average forgetting across all learning units is given by:

$$AF = \frac{1}{U-1}\sum_{u=1}^{u-1} F_u \quad (9)$$

Lower AF values indicate better knowledge retention, with a zero value implying no forgetting.

### B. Localization Error Comparisons with State-of-the-art

Fig. 5 shows the mean localization error for all comparison frameworks in Building 1 across devices and regions, under three CL scenarios: DIL-Exclusive, CIL-Exclusive, and CDIL with whiskers due to device heterogeneity. For the DIL-Exclusive scenario in Building 1 (Fig. 5(a)), DAILOC, which disentangles RSS fingerprints into domain-invariant and domain-specific latent spaces, shows reduced device variance with mean error of 1.15 meters. CIELO achieves a slightly lower mean error of 0.99 meters but higher device variance due to its replay-based strategy. MoCo-ON, designed with an orthogonal negative contrastive loss to bridge domain shifts, performs moderately at 1.39 meters. MOELO, with its device-invariant encoder and ETF-based MoE routing, delivers the lowest mean localization error of 0.96 meters.

In the CIL-Exclusive problem scenario for Building 1 (Fig. 5(b)), where whiskers vanish as all devices are always present during each training step, MOELO outperforms all other frameworks with the lowest mean localization error of 1.19 meters, benefiting from its region-specialized experts. In contrast, error for CIELO, MoCo-ON, and DAILOC degrades to 3.63 meters, 10.12 meters, and 16.87 meters, respectively, reflecting the absence of robust CIL mechanisms.

The CDIL problem scenario for Building 1 (Fig. 5(c)), is most revealing as tailored for combined class and domain shifts, MoCo-ON performs moderately at 4.33 meters, while CIELO (13.37 meters) and DAILOC (18.50 meters) collapse under compounded shifts with large device variability. MOELO again stands out with the lowest mean localization error of 0.52 meters, benefiting from its modular MoE experts and robust ETF gating, while also maintaining consistently low device variability. An interesting observation here is that CDIL scenarios show lower mean error than CIL scenarios, as incrementally learning devices allows better adaptation than requiring models to generalize across all devices at once.

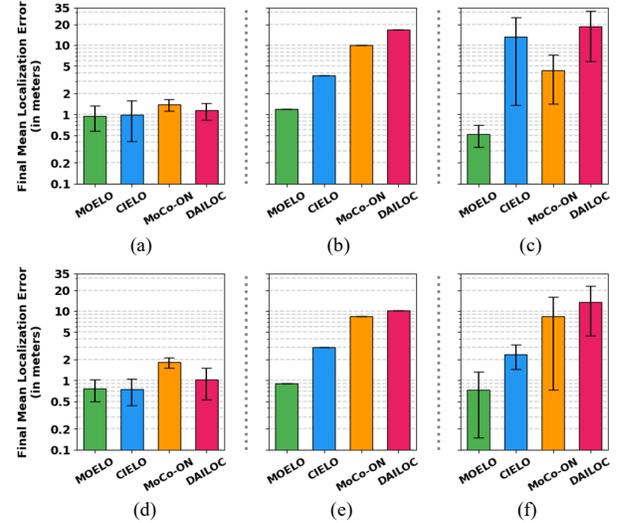

Fig. 5. Continual learning scenarios across two buildings. Subfigures (a–c) illustrate analysis in Building 1 with (a) DIL-Exclusive, (b) CIL-Exclusive, and (c) CDIL scenarios. Subfigures (d–f) illustrate analysis in Building 2 with: (d) DIL-Exclusive, (e) CIL-Exclusive, and (f) CDIL scenarios.

In Building 2 (Fig. 5(d)–(f)), MOELO again delivers the best performance, recording the lowest errors in DIL (0.77 meters), CIL (0.92 meters), and CDIL (0.74 meters) scenarios. CIELO remains competitive in DIL (0.75 meters) but its replay-based strategy falters under cross-domain increments, with errors rising to 3.03 meters in CIL and 2.38 meters in CDIL. MoCo-ON manages moderate accuracy in DIL (1.83 meters) but collapses under CIL (8.54 meters) and CDIL (8.42 meters), showing its contrastive alignment alone cannot adapt to class shifts. DAILOC performs reasonably in DIL (1.02 meters) yet deteriorates sharply when new regions and devices are introduced, with errors of 10.34 meters in CIL and 13.68 meters in CDIL, as disentanglement helps with adaptation to devices but not with learning new regions. These results suggest that Building 2, with its different RP and AP distribution, magnifies the weakness of domain-only or class-only strategies, while highlighting MOELO's robustness in handling both domain and class shifts.

*In summary*, in DIL-Exclusive scenarios, MOELO outperforms MoCo-ON by 1.9× and DAILOC by 1.3× while matching CIELO. In CIL-Exclusive scenarios, it improves upon DAILOC by 12.9×, MoCo-ON by 8.9×, and CIELO by 3.2×. In CDIL scenarios, MOELO achieves 25.6×, 12.5×, and 10.1× lower errors than DAILOC, CIELO, and MoCo-ON.

### C. Comparing Forgetting with State-of-the-art

A comparison of how each framework retains knowledge of previously learned devices and regions as learning progresses in Building 1 is shown in Fig. 6. The x-axis denoted incremental learning instances, with annotations below each step indicating the specific increment (device or region) introduced. The y-axis reports the mean localization error on all previously seen data, where lower values and flatter curves indicate stronger retention and reduced forgetting. The marker shapes further indicate the type of increment: circles for DIL, squares for CIL, and triangles for CDIL. In the DIL-Exclusive scenario (Fig. 6(a)), MOELO maintains stable performance, while CIELO weakens under device shifts. In the CIL-Exclusive scenario (Fig. 6(b)), forgetting is more severe for CIELO and DAILOC, whereas MOELO's modular expert design preserves prior region performance. Under the CDIL scenario (Fig. 6(c)), MoCo-ON performs moderately but still tends to forget, while MOELO sustains robustness.

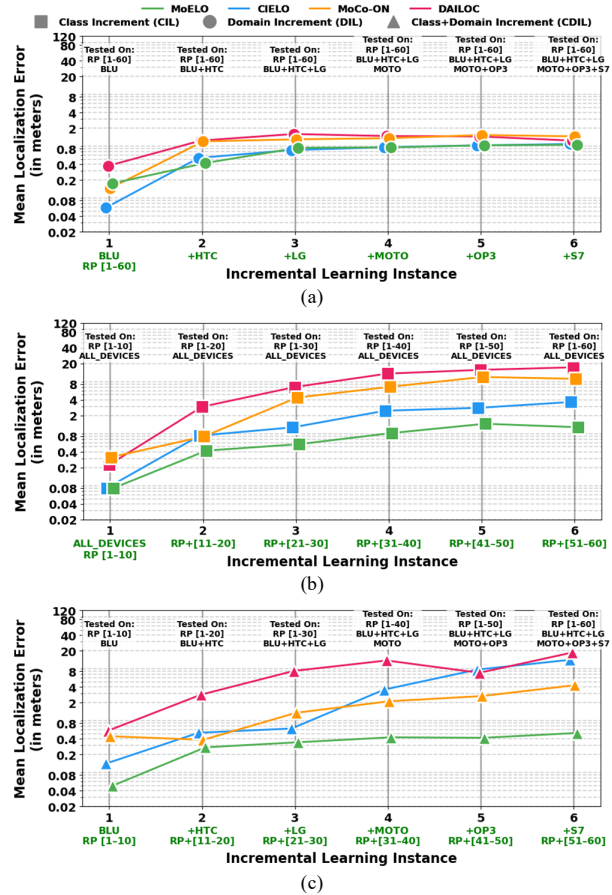

Fig. 6. Comparing forgetting as learning progresses in Building 1 : (a) DIL-Exclusive, (b) CIL-Exclusive, and (c) CDIL scenarios.

Quantitatively, MOELO achieves the best stability in all scenarios with an AF score of 0.26 (1.0×) compared to CIELO at 3.34 (12.8× worse), MoCo-ON at 3.24 (12.45× worse), and DAILOC at 5.61 (21.5× worse). Although results for Building 2 are not shown due to lack of space, a similar pattern is observed: compared to MOELO with an AF score of 0.23, CIELO exhibits an AF of 0.72 (3× worse), MoCo-ON 3.41 (14.3× worse), and DAILOC 3.85 (16.1× worse), reinforcing MOELO's robustness across diverse buildings.

Overall, MOELO clearly stands out by combining low localization error with minimal forgetting across all scenarios, outperforming the state-of-the-art.

### D. Comparing Model Efficiency

Table III compares model size and inference latency for all frameworks, measured on an HTC U11 with Qualcomm Snapdragon 835 processor. MOELO, despite being the most modular with its MoE design, maintains the smallest parameter count (86,904) and only slightly higher inference latency (0.83 ms) compared to others. This reflects the efficiency of ETF-based gating, which activates only one expert at inference, keeping runtime overhead low. CIELO is also lightweight, but less robust to domain shifts. MoCo-ON and DAILOC incur larger footprints due to their momentum encoder and disentanglement modules, respectively. Overall, MOELO offers compact size and competitive latency, making it well-suited for smartphone deployment.

TABLE III. MODEL PARAMETER COUNT AND LATENCY COMPARISON

| Framework | Total Parameters | Inference Latency |
|---|---|---|
| **MOELO** | 86,904 | 83 ms |
| **CIELO** | 89,405 | 62 ms |
| **MoCo-ON** | 109,692 | 71 ms |
| **DAILOC** | 278,984 | 69 ms |

### E. Sensitivity Analysis: Region Granularity

Fig. 7 presents a sensitivity analysis on Building 1 with respect to region granularity, where the total number of regions is varied while keeping the overall number of RPs fixed. This effectively changes the number of RPs assigned per region ($n_{RP}$), ranging from coarse partitioning (few regions, large $n_{RP}$) to fine partitioning (many regions, small $n_{RP}$). Across all frameworks, errors remain low when the number of regions is small (large $n_{RP}$). However, as the region count increases and $n_{RP}$ drops below 15, CIELO, MoCo-ON, and DAILOC show sharp error increase, with DAILOC degrading the most under finer splits. In contrast, MOELO remains stable, consistently keeping error below 0.57 meters, showing superior robustness to region granularity. For brevity, results for Building 2 are not shown, but similar trends were observed. While other frameworks become unstable with finer partitions, MOELO sustains a localization error below 0.45 meters, showing its robustness across buildings.

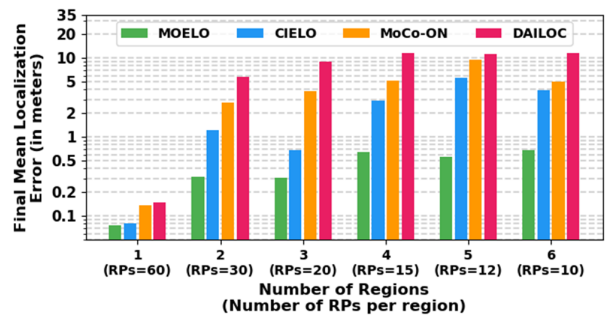

Fig. 7. Comparing sensitivity to region granularity on Building 1.

## VI. Conclusion

This paper introduced MOELO, the first unified class and domain incremental continual learning framework for Wi-Fi fingerprinting based indoor localization. Unlike prior approaches that address class-incremental or domain-incremental learning separately, MOELO integrates both through a mixture-of-experts design with ETF-based gating. Expert specialization, stable routing, and balanced replay together enable long-term adaptability without catastrophic forgetting. Experimental evaluations across buildings and mobile devices demonstrate that MOELO delivers substantial gains over state-of-the-art frameworks, achieving up to $25.6\times$ lower mean localization error, $44.5\times$ lower worst-case error, and $21.5\times$ less forgetting. These improvements are realized while maintaining a lightweight model footprint, ensuring suitability for real-world smartphone deployments.